\documentclass[conference]{IEEEtran}

\usepackage{cite}
\usepackage{amsmath,amssymb,amsfonts}
\usepackage{textcomp}

\usepackage{graphicx}

\usepackage{xcolor}

\usepackage{multirow}
\usepackage{booktabs}

\usepackage{enumitem}
\setlist[itemize]{leftmargin=*, topsep=.0em, itemsep=0pt, parsep=0pt, partopsep=0pt}
\setlist[enumerate]{leftmargin=*, topsep=.0em, itemsep=0pt, parsep=0pt, partopsep=0pt}

\usepackage[normalem]{ulem}
\useunder{\uline}{\ul}{}

\usepackage{algorithm}
\usepackage[noend]{algpseudocode}

\newcommand*\Input[1]{\Statex \textbf{Input:} #1}
\newcommand*\Output[1]{\Statex \textbf{Output:} #1}
\algrenewcommand\alglinenumber[1]{#1}

\newcommand{\forceindent}{\leavevmode{\parindent=1em\indent}}

\usepackage{amsthm}
\newtheoremstyle{def_style}
  {.0em}      
  {.0em}      
  {}          
  {}          
  {\bfseries} 
  {.}         
  {.0em}      
  {}          

\theoremstyle{def_style}

\theoremstyle{def_style}

\usepackage{caption}
\usepackage{subcaption}

\usepackage[hyphens]{url}

\usepackage{amsmath}
\usepackage{bbm}

\usepackage{collectbox}

\makeatletter
\newcommand{\mybox}{%
    \collectbox{%
        \setlength{\fboxsep}{1pt}%
        \fbox{\BOXCONTENT}%
    }%
}
\makeatother

\begin{document}

\title{Ego-Network Transformer for Subsequence Classification in Time Series Data}



\author{\IEEEauthorblockN{Chin-Chia Michael Yeh, Huiyuan Chen, Yujie Fan, Xin Dai, Yan Zheng, \\Vivian Lai, Junpeng Wang, Zhongfang Zhuang, Liang Wang, Wei Zhang, Eamonn Keogh$^\dagger$}
\IEEEauthorblockA{\textit{Visa Research}, \textit{University of California, Riverside}$^\dagger$ \\
\{miyeh,hchen,yufan,xidai,yazheng,viv.lai,junpenwa,zzhuang,liawang,wzhan\}@visa.com}
}

\maketitle

\begin{abstract}
Time series classification is a widely studied problem in the field of time series data mining. 
Previous research has predominantly focused on scenarios where relevant or \textit{foreground} subsequences have already been extracted, with each subsequence corresponding to a single label. 
However, real-world time series data often contain foreground subsequences that are intertwined with \textit{background} subsequences. 
Successfully classifying these relevant subsequences requires not only distinguishing between different classes but also accurately identifying the foreground subsequences amidst the background.
To address this challenge, we propose a novel subsequence classification method that represents each subsequence as an ego-network, providing crucial nearest neighbor information to the model. 
The ego-networks of all subsequences collectively form a time series subsequence graph, and we introduce an algorithm to efficiently construct this graph. 
Furthermore, we have demonstrated the significance of enforcing temporal consistency in the prediction of adjacent subsequences for the subsequence classification problem.
To evaluate the effectiveness of our approach, we conducted experiments using 128 univariate and 30 multivariate time series datasets. 
The experimental results demonstrate the superior performance of our method compared to alternative approaches.
Specifically, our method outperforms the baseline on 104 out of 158 datasets.
\end{abstract}






\section{Introduction}
\begin{figure}[ht]
\centerline{
\includegraphics[width=0.9\linewidth]{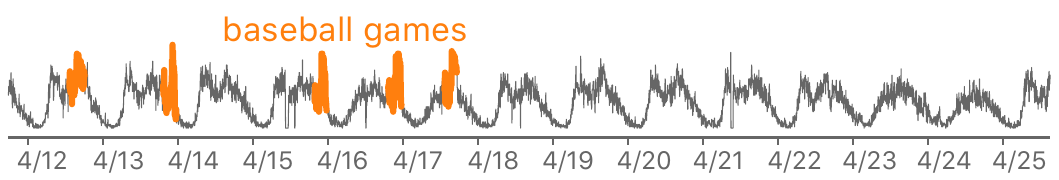}
}
\caption{
The traffic time series data is from highway around Dodger Stadium in Los Angeles, California~\cite{ihler2006adaptive}.
}
\label{fig:dodgers}
\end{figure}

The time series classification problem is widely studied in the data mining community, with numerous approaches proposed to classify segmented time series into their respective classes~\cite{ruiz2021great}.
However, in many real-world scenarios, time series can contain multiple relevant foreground classes mixed with irrelevant background segments.
For example, consider a time series obtained by monitoring highway traffic, which predominantly consists of regular traffic but occasionally includes more interesting events such as baseball games, road works, or car crashes.
The traffic events caused by baseball games are highlighted in Fig.~\ref{fig:dodgers}.

One approach to address this problem is to build a classifier that focuses on classifying subsequences into foreground or background classes based on the training data.
During testing, the model continuously classifies incoming subsequences.
As we perform classification on subsequences, we refer to this problem as the \textit{subsequence classification problem}, which is the main focus of this paper.

However, when applying conventional time series classification models to this problem, we have observed cases where even sophisticated neural network models are outperformed by simple $k$-nearest neighbor classifiers using $z$-normalized Euclidean distance.
To address this challenge, we propose an \textit{ego-network Transformer model} that combines the strengths of both approaches.
The model learns representations of different subsequences using neural network models and integrates them with a Transformer model based on ego-networks extracted from $k$-nearest neighbor graphs.
Furthermore, constructing $k$-nearest neighbor graphs can be computationally intensive when considering every subsequence in the target time series.
To overcome this, we introduce an efficient $k$-nearest neighbor graph algorithm for both training and test cases.

An additional benefit of our ego-network Transformer design is its improved efficiency over a naive subsequence-based Transformer.
A simple implementation of the Transformer for time series subsequence would involve each subsequence attending to every other subsequence in the time series.
However, this approach is inefficient in terms of both time and space complexity.
For instance, if there are $n$ subsequences in the time series, it would require $O(n^2)$ for training the model and $O(n)$ for inference.
In contrast, our ego-network Transformer model only attends to its $k$ nearest neighbors ($k \ll n$), resulting in training and testing complexities of $O(k^2)$ and $O(k)$, respectively.

Another crucial aspect of subsequence classification is temporal consistency. 
Due to the significant overlap between these subsequences, it is essential to ensure that the predictions for each subsequence are consistent and aligned with their temporal context.
To leverage the benefits of temporal consistency, we have developed a simple yet effective post-processing technique. 
This technique involves comparing the predicted labels of adjacent subsequences within their temporal context, with the aim of reducing false positives and false negatives. 
By incorporating this post-processing step, we are able to enhance the overall performance of the classification method.

The contributions of this paper are:

\begin{itemize}
    \item We propose an ego-network Transformer model that combines the strengths of conventional time series classification models and $k$-nearest neighbor classifiers. 
    The model integrates representations learned by neural network models with a Transformer model based on ego-networks extracted from $k$-nearest neighbor graphs.
    \item We introduce an efficient algorithm for constructing $k$-nearest neighbor graphs, alleviating the computational burden associated with considering every subsequence in the target time series.
    \item We develop a post-processing technique to enforce temporal consistency in the predictions of adjacent subsequences.
    \item Through extensive experiments on 128 univariate time series datasets and 30 multivariate time series datasets, we demonstrate the superior performance of our proposed ego-network Transformer model compared to baseline models.
    \item We conduct a case study to validate the importance of the nearest neighbor graph in subsequence classification, particularly in datasets with data scarcity issues.
\end{itemize}


\section{Background}
\label{sec:background}
In this section, we will begin by presenting the problem statement for the subsequence classification task by highlighting its distinctions from other types of time series classification problems.
Following that, we will conduct a review of related works, exploring the existing approaches in the field.

\subsection{Problem Statement}
\begin{figure}[ht]
\centerline{
\includegraphics[width=0.85\linewidth]{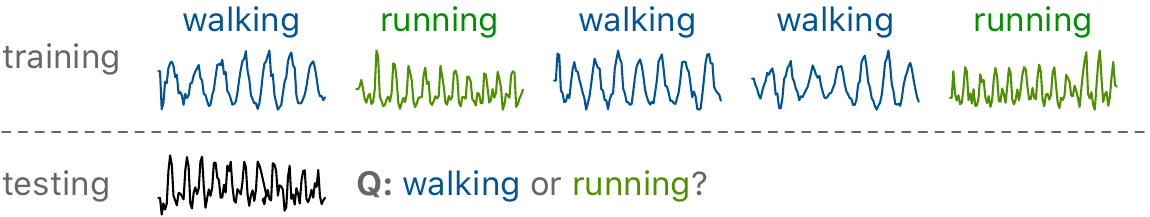}
}
\caption{
The time series classification problem.
}
\label{fig:problem_0}
\end{figure}

Extensive research has been conducted on various classification problems involving time series~\cite{abdoli2018time,bagnall2018uea,dau2019ucr,li2019segmentation}.
One commonly studied variant is known as \textit{time series classification}~\cite{dau2019ucr,bagnall2018uea}.
An example of such problem is shown in Fig.~\ref{fig:problem_0}.

During the training phase, a machine learning model is trained using a set of training time series, along with their corresponding ground truth labels.
In this particular example, the time series data consists of human activity recordings obtained from accelerometers.
Each time series is assigned a label indicating whether the activity is classified as \textit{walking} or \textit{running}.
Once a test time series $X$ is obtained, the model predicts the most likely class label for $X$.

\begin{figure}[ht]
\centerline{
\includegraphics[width=0.9\linewidth]{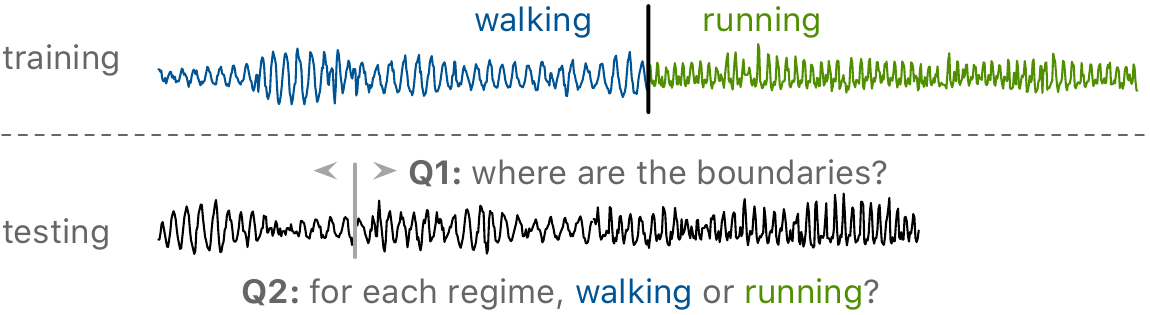}
}
\caption{
The semantic segmentation problem.
}
\label{fig:problem_1}
\end{figure}

Another closely related time series problem is known as \textit{semantic segmentation}~\cite{li2019segmentation}.
An example of a time series semantic segmentation problem involving human activity time series is depicted in Fig.~\ref{fig:problem_1}.

Unlike the time series classification problem, in semantic segmentation, we no longer have a dataset consisting of multiple time series.
Instead, we have a single training time series~$T_\text{train}$ and a separate testing time series~$T_\text{test}$.
The training time series~$T_\text{train}$ can be segmented into multiple regimes, and for the ground truth labels, we have the locations of the boundaries between the regimes and the corresponding ``class" of each regime.
During testing, the trained model needs to perform two tasks: 1) identify the segmentation boundaries and 2) classify each regime into the appropriate class.

\begin{figure}[ht]
\centerline{
\includegraphics[width=0.9\linewidth]{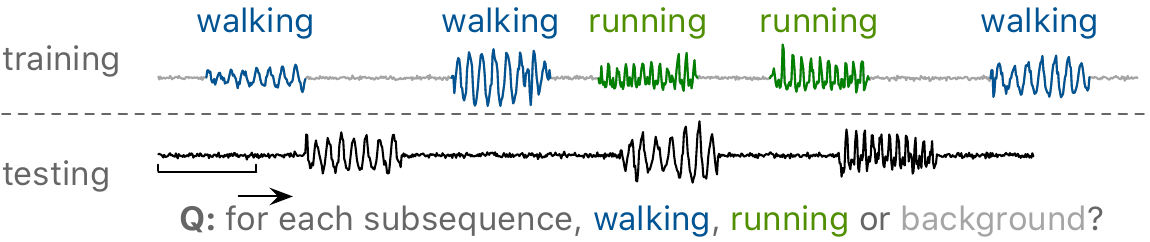}
}
\caption{
The subsequence classification problem.
}
\label{fig:problem_2}
\end{figure}

Finally, the problem addressed in this paper is referred to as the \textit{subsequence classification} problem, and an example of such a problem is illustrated in Fig.~\ref{fig:problem_2}.

Similar to semantic segmentation, the subsequence classification problem involves a single training time series~$T_\text{train}$ and a separate testing time series~$T_\text{test}$.
However, both $T_\text{train}$ and $T_\text{test}$ may contain background segments that are unrelated to the classes of interest.
In Fig.~\ref{fig:problem_2}, the relevant subsequences correspond to walking and running patterns, while the other subsequences are considered as background segments (e.g., standing still).
During testing, predictions are made at the subsequence level, where the subsequences are generated using a sliding window approach.
The machine learning model needs to determine whether a subsequence belongs to one of the relevant classes (e.g., walking or running) or if it is a background subsequence.

It is important to note that the example problems presented in this section focus on univariate time series. 
However, each of these problems can also be formulated and extended to handle multivariate time series data.

\subsection{Related Work}
We focus our related works section on two topics: 1) $k$-nearest neighbor subsequence graph and 2) time series classification.

The notion of a $k$-nearest neighbor subsequence graph may not have been explicitly explored in the time series data mining community. 
However, a similar concept has been implicitly adopted in the form of the Matrix Profile~\cite{yeh2016matrix,yeh2018time}.
The Matrix Profile algorithm involves computing two meta time series for a given time series $T$: the Matrix Profile and the Matrix Profile Index. 
The Matrix Profile stores the distance (typically $z$-normalized Euclidean distance) between each subsequence and its nearest neighbor, while the Matrix Profile Index stores the identity of the nearest neighbor. 
Together, these two meta time series form a \textit{1-nearest neighbor graph} for all subsequences in $T$. 
This 1-nearest neighbor subsequence graph has been utilized in various time series data mining tasks, including motif discovery, anomaly detection, and segmentation~\cite{athira2022comprehensive,anton2018time,kieu2019outlier,gharghabi2017matrix,ermshaus2023clasp}.

The initial Matrix Profile algorithms, such as STOMP~\cite{zhu2016matrix}, have already demonstrated sufficient efficiency for large-scale time series. 
However, subsequent research has made significant progress in further reducing the computational time of the Matrix Profile. 
Approaches such as utilizing specialized hardware, approximation techniques, and improved anytime convergence have been adopted to enhance its efficiency~\cite{fernandez2022tratsa,fernandez2020natsa,zimmerman2019matrix,yeh2022error,zhu2018matrix}. 
Many of these techniques can be incorporated into our $k$-nearest neighbor construction algorithm.
Nevertheless, as the first work in adopting the $k$-nearest neighbor graph for the subsequence classification problem, our primary focus is to demonstrate the benefits of using this graph in our proposed approach. 
Therefore, we have chosen to extend a more basic version of the Matrix Profile algorithm, such as STOMP~\cite{zhu2016matrix}, to avoid the additional complexity associated with these more advanced methods. 
Exploring the integration of these advanced techniques into our algorithm is an avenue for future work.

Over the years, numerous time series classification methods have been proposed~\cite{bagnall2017great,ruiz2021great}. 
Recent benchmark papers~\cite{bagnall2017great,ruiz2021great} have identified methods such as HIVE-COTE~\cite{middlehurst2021hive}, ROCKET~\cite{dempster2020rocket}, and ResNet~\cite{wang2017time} to achieve state-of-the-art performance in time series classification. 
In our work, we have chosen to extend neural network-based methods, e.g., ResNet~\cite{wang2017time}, due to their modular nature, which facilitates easy modification and integration into our approach as backbone models. 
The designs of our backbone time series representation learning models are inspired by various popular neural network architectures for modeling sequential data~\cite{hochreiter1997long,cho2014properties,vaswani2017attention,wang2017time,zhou2021informer}, as introduced in Section~\ref{sec:backbone}.
\newcommand{\figpatternzero}[1]{\includegraphics[height=1.1em]{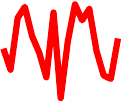}#1}
\newcommand{\figpatternone}[1]{\includegraphics[height=1.1em]{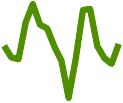}#1}
\newcommand{\figpatterntwo}[1]{\includegraphics[height=1.1em]{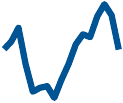}#1}

\section{Time Series Subsequence Graph}
\label{sec:ts_graph}
\begin{figure}[ht]
\centerline{
\includegraphics[width=0.85\linewidth]{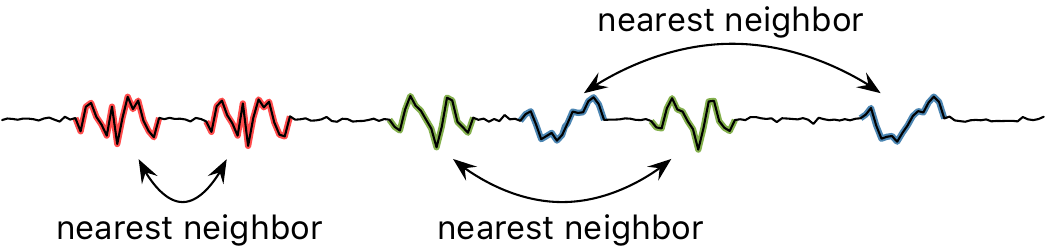}
}
\caption{
Users can identify highly preserved patterns using the 1-nearest neighbor graph.
}
\label{fig:1nn_ts}
\end{figure}

The time series subsequence graph captures relationships between subsequences of a given time series. 
Specifically, the graph aims to capture nearest neighbor relationships between subsequences based on similarity in shape. 
In Fig.~\ref{fig:1nn_ts}, the time series contains three pairs of embedded patterns: \figpatternzero{,} \figpatternone{,} and \figpatterntwo{.} 
By analyzing the most similar pairs of subsequences in the 1-nearest neighbor graph, these highly preserved patterns can be quickly identified.

One notable example of utilizing the time series subsequence graph in time series data mining is the Matrix Profile~\cite{yeh2018time,zhu2020swiss}. 
It proves to be effective in accomplishing various tasks, including motif discovery, anomaly detection, and semantic segmentation, by leveraging the power of 1-nearest neighbor subsequence graphs~\cite{yeh2018time,zhu2020swiss,yeh2022error}. 
In this paper, we capitalize on this versatile representation of subsequence relationships to address the subsequence classification problem. 
The proposed method leverages ego-networks for each subsequence extracted from the $k$-nearest neighbor subsequence graphs to enhance the performance of subsequence classification models.

The problem of constructing the $k$-nearest neighbor subsequence graph can be naively solved by extracting all subsequences and computing the pairwise distances between them.
However, this approach would result in a time complexity of $O(n^2m)$, where $n$ is the length of the time series and $m$ is the subsequence length.
To construct the $k$-nearest neighbor subsequence graph more efficiently, we leverage an extension of the STOMP algorithm~\cite{zhu2016matrix,yeh2018time}, originally designed for computing the Matrix Profile. 
This extension is possible because one interpretation of the Matrix Profile is that it represents the 1-nearest neighbor graph for all subsequences in the time series~\cite{yeh2016matrix,yeh2018time}. 
In other words, we are expanding the STOMP algorithm from a 1-nearest neighbor graph to a $k$-nearest neighbor graph.
The pseudo code for the extended STOMP algorithm is presented in Algorithm~\ref{alg:knn_stomp}.

\begin{algorithm}[ht]
    \centering
    \caption{$k$-Nearest Neighbor STOMP Algorithm\label{alg:knn_stomp}}
    \footnotesize
    \begin{algorithmic}[1]
        \Input{time series~$T \in \mathbb{R}^n$, subsequence length~$m \in \mathbb{N}$, number of neighbors~$k \in \mathbb{N}$}
        \Output{$k$-nearest neighbor index~$\mathbf{I} \in \mathbb{N}^{(n - m + 1) \times k}$}
        \Function{$k$NNSTOMP}{$T, m, k$}
        \State $\mathbf{I} \gets $ zero matrix with size $(n - m + 1) \times k$
        \For{$i \in [0, \cdots, n - m + 1]$}
        \State $Q \gets T[i:i+m]$
        \State $D \gets$ \textsc{GetDistanceProfile}$(Q, T)$
        \State $D \gets$ \textsc{MaskingWithInf}$(D, i)$
        \For{$j \in [0, \cdots, k]$}
        \State $\mathbf{I}[i,j] \gets$\textsc{FindMinIndex}$(D)$
        \State $D \gets$ \textsc{MaskingWithInf}$(D, \mathbf{I}[i,j])$
        \EndFor
        \EndFor
        \State \Return $\mathbf{I}$
        \EndFunction
    \end{algorithmic}
\end{algorithm}

Algorithm~\ref{alg:knn_stomp} demonstrates the construction process of the $k$-nearest neighbor graph for the training data. 
In the later paragraph, we will discuss the necessary modifications to adapt Algorithm~\ref{alg:knn_stomp} for the testing scenario. 
Algorithm~\ref{alg:knn_stomp} takes the training time series $T \in \mathbb{R}^n$, the subsequence length $m \in \mathbb{N}$, and the number of neighbors $k \in \mathbb{N}$ as input. 
The length of $\mathbf{T}$ is denoted by $n$. 
The algorithm outputs a matrix that stores the index of the $k$-nearest neighbor for each subsequence in~$T$.

In line 2, we initialize a matrix $\textbf{I}$ to store the indices of the $k$-nearest neighbors.
The number of subsequences in $T$ is $n - m + 1$.
The for loop from line 3 to line 10 iterates through each subsequence to find its $k$-nearest neighbors.
In line 4, we extract the query subsequence $Q$.
In line 5, we compute the distance profile between $Q$ and $T$.
The distance profile $D \in \mathbb{R}^{n - m + 1}$ stores the $z$-normalized Euclidean distance between $Q$ and each subsequence in $T$.
For instance, $D[i]$ stores the distance between $Q$ and $T[i:i+m]$.
The naive implementation of this step has a time complexity of $O(nm)$. However, utilizing the technique presented in~\cite{zhu2016matrix, yeh2018time}, the time complexity can be reduced to $O(n)$.

In line 6, an exclusion zone is applied to the distance profile $D$ to avoid \textit{trivial matches} with the query subsequence $Q$. 
A trivial match occurs when the nearest neighbor of $Q$ in $T$ is $Q$ itself~\cite{mueen2009exact}. 
This situation arises when $Q$ is a subsequence of $T$.
By definition, if $Q$ is the $i$th subsequence of $T$, $D[i]$ is zero, and the values around the $i$th position in $D$ would be very close to zero. 
To prevent these subsequences from being considered as nearest neighbors, we replace the values around the $i$th position in $D$ with infinity. 
In our implementation, if the input index to the \textsc{MaskingWithInf}$()$ function is $i$, we set $D[i - \frac{m}{2}:i + \frac{m}{2}]$ to infinity.

From line 7 to line 9, the $k$ neighbors of $Q$ in $T$ are identified using $D$.
In line 8, the nearest neighbor is determined by finding the index of the minimal value in $D$.
In line 9, the same \textsc{MaskingWithInf}$()$ function is applied to prevent the same nearest neighbor from being found in the next iteration.
The output is returned in line 10.
The time complexity of the algorithm is $O(n^2)$, as it involves computing $n - m + 1$ distance profiles, and the space complexity is $O(kn)$ for storing the output matrix $\mathbf{I}$.

Table~\ref{tab:knn_time} presents the runtime of various $k$-nearest neighbor graph construction algorithms for different input time series lengths.
The \textit{Naive} algorithm refers to the brute force implementation, while \textit{STAMP-based} is an extension of the STAMP algorithm~\cite{yeh2016matrix}.
The algorithm adopted in this paper is referred to as \textit{STOMP-based}.
Both the STAMP-based and STOMP-based algorithms exhibit significantly improved efficiency compared to the naive implementation, with the STOMP-based algorithm being the most efficient among them.

\begin{table}[ht]
\caption{Runtime of different graph construction algorithms in seconds.
The first row contains the length of input time series.
}
\begin{center}
\resizebox{0.99\columnwidth}{!}{
\begin{tabular}{l||cccccc}
runtime ($\downarrow$) & 500 & 1,000 & 1,500 & 2,000 & 2,500 & 3,000 \\ \hline \hline
Naive & 16.24 & 89.02 & 208.22 & 386.49 & 612.32 & 907.03 \\
STAMP-based & 0.0653 & 0.2102 & 0.4278 & 0.7082 & 1.0878 & 1.5323 \\
STOMP-based & 0.0357 & 0.0921 & 0.1559 & 0.2356 & 0.3147 & 0.4393  \\
\end{tabular}
}
\label{tab:knn_time}
\end{center}
\end{table}

In the testing scenario, we have two time series: the training time series $T_\text{train} \in \mathbb{R}^{n_\text{train}}$ and the test time series $T_\text{test} \in \mathbb{R}^{n_\text{test}}$.
Since the objective is to construct the ego-network for each subsequence in $T_\text{test}$ with respect to the subsequences in $T_\text{train}$, the output matrix $I$ will have a size of $(n_\text{test} - m + 1) \times k$.
To accommodate this change, line 2 in Algorithm~\ref{alg:knn_stomp} needs to be modified accordingly.
In line 3, the range of $i$ is adjusted to $[0, \cdots, n_\text{test} - m + 1]$.
Line 4 and line 5 are modified as $Q \gets T_\text{test}[i:i+m]$ and $D \gets$ \textsc{GetDistanceProfile}$(Q, T_\text{train})$ respectively.
Since $Q$ is not a subsequence of $T_\text{train}$, there is no trivial match problem, and thus line 6 is removed.
Lines 7 to 10 remain unchanged for the testing scenario.
The modified version of the algorithm has a time complexity of $O(n_\text{train} n_\text{test})$ and a space complexity of $O(\textsc{Max}(n_\text{train}, k n_\text{test}))$.

Although the SCAMP algorithm~\cite{zimmerman2019matrix} offers the potential for achieving even higher efficiency, we decided against adopting it. 
The reason is that the algorithm constructs an approximated $k$-nearest neighbor subsequence graph, and the impact of this approximation on the final classification accuracy remains unknown. 
While incorporating the SCAMP algorithm for improved efficiency would be an intriguing extension to our current system, we have left it for future exploration.

When working with multidimensional time series, we calculate the $z$-normalized Euclidean distance using all dimensions. 
Following a similar approach as in~\cite{yeh2017matrix}, for a given pair of multidimensional subsequences, we compute the $z$-normalized Euclidean distance between them for each dimension and then aggregate the distances across different dimensions by summing them. 
In essence, each dimension contributes equally to the distance between the subsequences. 
It would be interesting to investigate the concept of sub-dimensional nearest neighbors, as presented in~\cite{yeh2017matrix}, as we anticipate that sub-dimensional nearest neighbors would likely hold more meaningful comparisons than all-dimensional nearest neighbors. 
However, since the primary goal of this paper is to demonstrate the efficacy of the $k$-nearest neighbor graph in addressing the subsequence classification problem, we have deferred this extension for future research.

\section{Models and Methodology}
\label{sec:model}
In this section, we begin by introducing the proposed method for classifying an input subsequence using ego-network.
Next, we present the backbone models utilized to extract representations from the input time series data.
Then, we describe the training and inference algorithm associated with the proposed model.
Lastly, we discuss a simple yet effective post-processing technique designed to enhance the temporal consistency in the prediction of adjacent subsequences.

\subsection{Ego-Network Transformer Model}
\label{sec:ego_model}
\begin{figure}[ht]
\centerline{
\includegraphics[width=0.85\linewidth]{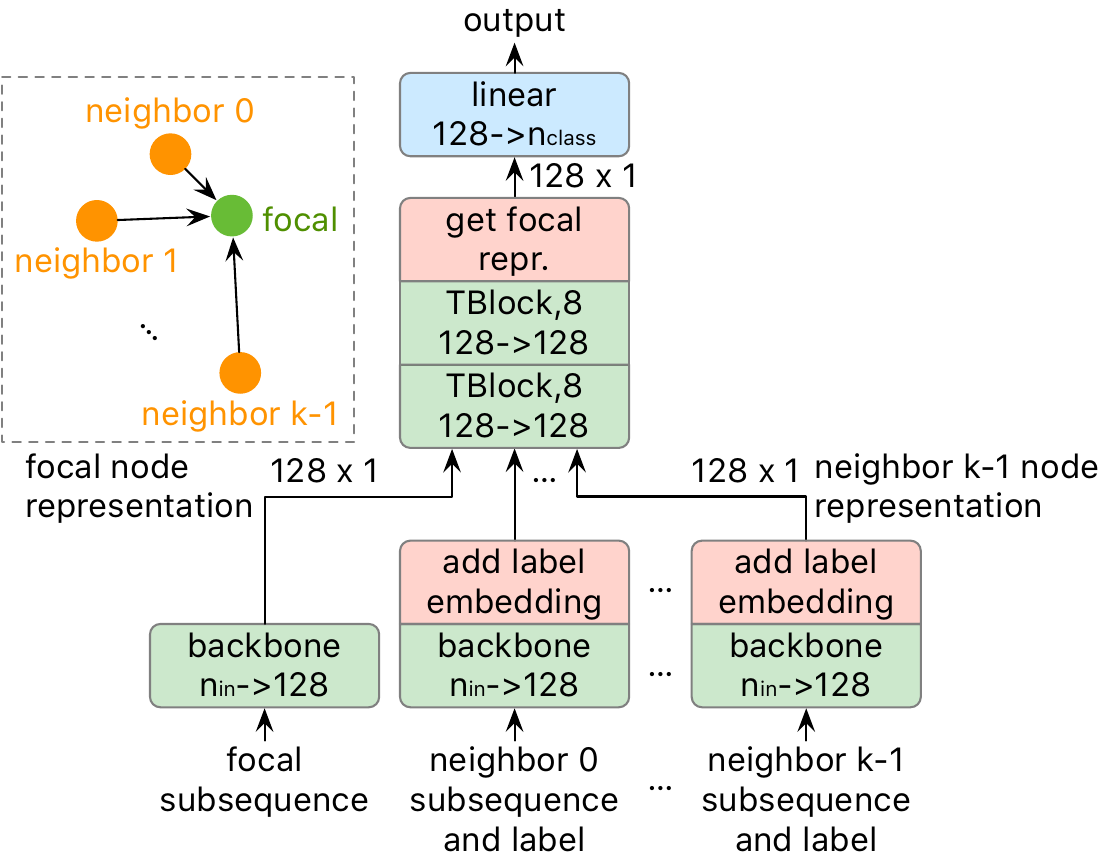}
}
\caption{
The proposed ego-network Transformer subsequence classification model.
}
\label{fig:egonet}
\end{figure}

The proposed Transformer model is presented in Fig.~\ref{fig:egonet}. 
The input to the model consists of a focal subsequence denoted as $X_\text{focal}$, along with its nearest neighbor subsequences from the training time series $X_0, \cdots, X_{k-1}$, and the corresponding labels of the neighbors $y_{0}, \cdots, y_{k-1}$.
Essentially, the inputs consist of the subsequences associated with the ego-network of the focal subsequence, where the ego-network is extracted from the $k$-nearest neighbor subsequence graph.
The initial step involves extracting the intermediate representation of each subsequence using one of the backbone models discussed in Section~\ref{sec:backbone}. 
These representations are denoted as $H_\text{focal}, H_0, H_1, \cdots, H_{k-1}$, where $H_\text{focal}$ corresponds to the representation of $X_\text{focal}$, and $H_0, H_1, \cdots, H_{k-1}$ correspond to the representations of $X_0, X_1, \cdots, X_{k-1}$, respectively.

Next, we incorporate the corresponding label for each neighbor subsequence by adding a learnable label embedding to its representation. 
Let $Y_i$ represent the label embedding for $y_i$, the label of the $i$th neighbor. 
The final subsequence representation for the neighbor is computed as $\hat{H}_i \gets Y_i + H_i$.
With the node representations of each subsequence prepared, we concatenate them together to form a set that includes all the node representations: $[H_\text{focal}, \hat{H}_0, \cdots, \hat{H}_{k-1}]$.
Subsequently, we employ two layers of Transformer blocks (refer to Fig.~\ref{fig:tblock}) to aggregate information from each node in the set.
After the Transformer blocks, we extract the representation corresponding to the focal subsequence. 
Finally, a linear layer is used to compute the logit for each class.
We choose to use a Transformer-based model design to capture the ego-network, as opposed to other graph neural networks like GCN~\cite{kipf2016semi} or GAT~\cite{velivckovic2017graph}, because it has been shown in~\cite{ying2021transformers} that Transformers are more effective compared to the alternatives.

The proposed method leverages the versatile and powerful $k$-nearest neighbor subsequence graph, as discussed in Section~\ref{sec:ts_graph}, for the subsequence classification problem. 
This approach offers notable advantages over attending to all subsequences in the training data. 
By focusing only on the top $k$ nearest neighbors, the method achieves improved efficiency. 
This is particularly significant considering the space complexity of the Transformer block, which grows quadratically with the number of input items. 
For instance, if the training data consists of one million subsequences, storing the attention matrix alone would require over seven terabytes of memory.

\subsection{Backbone Temporal Model}
\label{sec:backbone}
\begin{figure}[ht]
\centerline{
\includegraphics[width=0.75\linewidth]{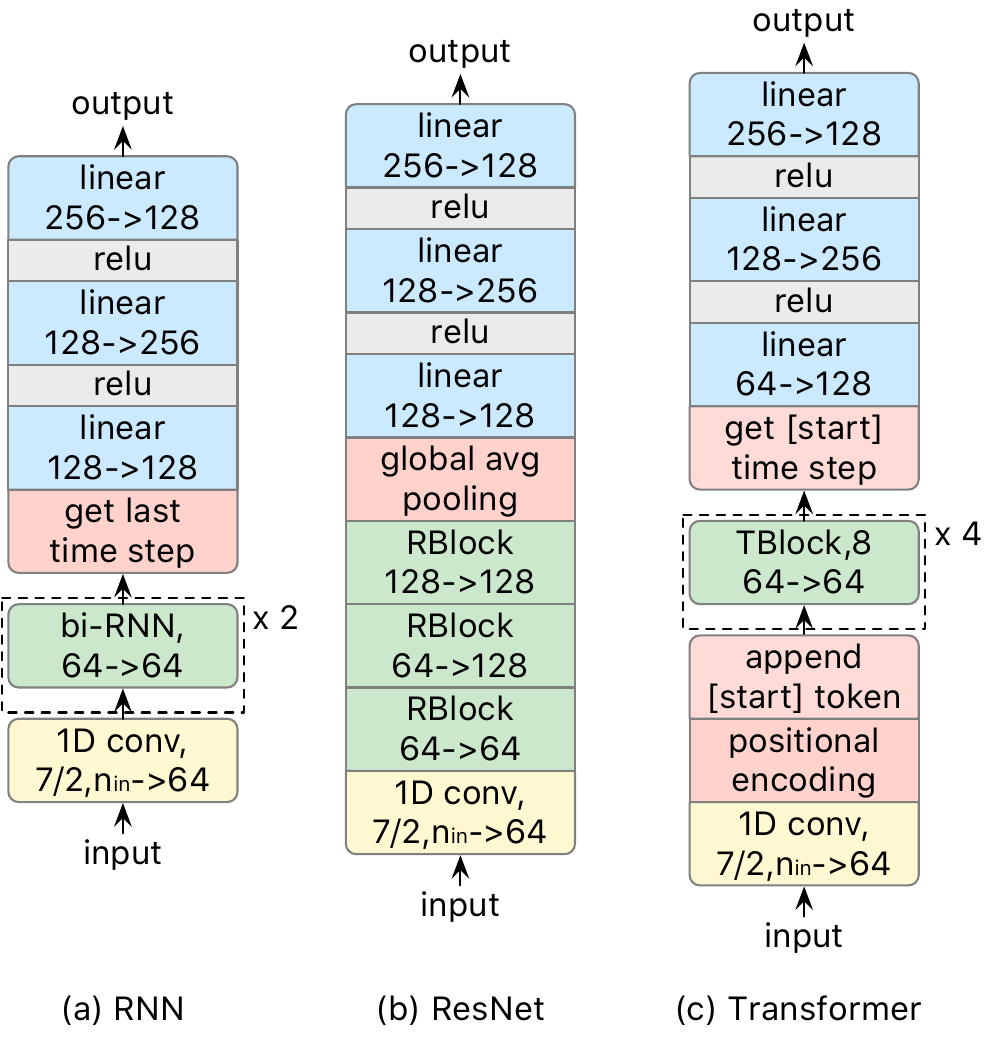}
}
\caption{
The designs of the backbone models are based on RNN, ResNet, and Transformer.
Please refer to Fig.~\ref{fig:rblock} for details about the \protect\mybox{\texttt{RBlock}} and Fig.~\ref{fig:tblock} for details about the \protect\mybox{\texttt{TBlock}}. 
}
\label{fig:backbone}
\end{figure}

\begin{figure}[ht]
\centerline{
\includegraphics[width=0.55\linewidth]{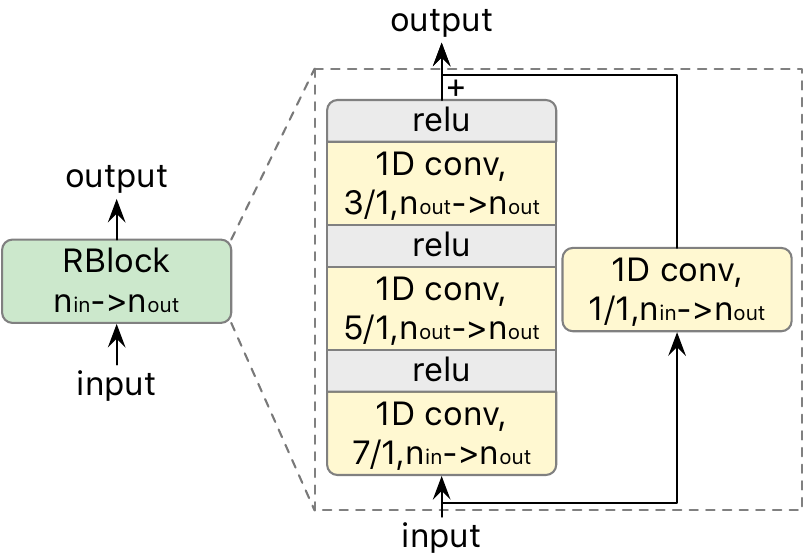}
}
\caption{
The designs of the residual block.
}
\label{fig:rblock}
\end{figure}

\begin{figure}[ht]
\centerline{
\includegraphics[width=0.4\linewidth]{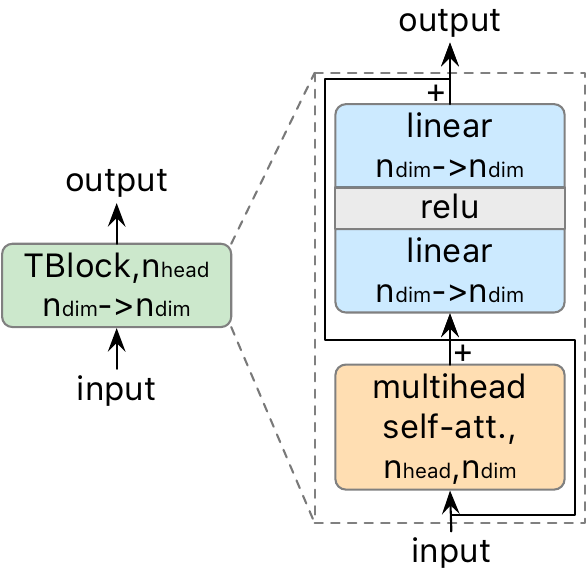}
}
\caption{
The designs of the Transformer block.
}
\label{fig:tblock}
\end{figure}

We have explored four different neural network architectures as backbone models for extracting global representations from time series data. 
A global representation captures the information from the entire input time series. 
The four backbone models are:

\begin{itemize}
\item The \textbf{Long Short-Term Memory Network (LSTM)} is a widely used type of Recurrent Neural Network (RNN) for modeling sequential data~\cite{hochreiter1997long,lim2021time,zhou2021informer}.
In our work, we adopt the design depicted in Fig.~\ref{fig:backbone}.a.
The figure employs specific notations to describe different layers. 
For instance, \mybox{\texttt{1D conv,7/2,$n_\text{in}{\to}$64}} represents a $1D$ convolutional layer with a filter size of 7, a stride size of 2, an input dimension of $n_\text{in}$, and an output dimension of 64.
Similarly, \mybox{\texttt{bi-RNN,64${\to}$64}} denotes a bidirectional RNN layer with an input dimension of 64 and an output dimension of 64.
In our case, the two bi-RNN layers are implemented as bidirectional LSTM layers.
Additionally, \mybox{\texttt{linear,64${\to}$64}} denotes a linear layer with an input dimension of 64 and an output dimension of 64.

\forceindent The input time series is first processed by the $1D$ convolutional layer to extract local patterns.
The decision to select only the last time step is based on the understanding that it encapsulates the information from the entire input time series. 
However, it is worth noting that the first time step could also be chosen since the RNN layers are bidirectional. 
In the end, the output of the LSTM backbone model consists of a size 128 vector for each input time series.

\item The \textbf{Gated Recurrent Unit Network (GRU)} is another popular type of RNN architecture for modeling sequential data~\cite{cho2014properties,lim2021time,zhou2021informer}.
We employ an identical design to the LSTM backbone model (see Fig.~\ref{fig:backbone}.a), with the only difference being that the two bi-RNN layers are implemented as GRU layers instead of LSTM layers.

\item The \textbf{Residual Network (ResNet)} is a time series classification model inspired by the success of ResNet in computer vision~\cite{he2016deep, wang2017time}. 
Extensive evaluations reported in~\cite{ismail2019deep} have demonstrated that ResNet is one of the most effective models for time series classification. 
Our design, depicted in Fig.~\ref{fig:backbone}.b, is based on the architecture proposed in~\cite{wang2017time}. 
In our notation, \mybox{\texttt{RBlock,64$\to$64}} represents a residual block (refer to Fig.~\ref{fig:rblock}) with an input dimension of 64 and an output dimension of 64.

\forceindent The length of the output sequence from the residual blocks depends on the length of the input time series. 
When the sequence length is greater than one, we employ a global average pooling function to generate a global intermediate representation of the input time series. 
The output of this backbone model is a size 128 vector that represents each input time series.

\item The \textbf{Transformer} is a widely adopted alternative to RNNs for sequence modeling~\cite{vaswani2017attention, li2019enhancing, zhou2021informer, lim2021time, chen2022denoising}.
In our work, we adopt the architecture depicted in Fig.~\ref{fig:backbone}.c. 
Like the previously discussed backbone models, the initial layer comprises a $1D$ convolutional layer designed to capture local patterns. 
To incorporate positional information, we follow~\cite{vaswani2017attention} and apply fixed positional encoding.
This encoding is added to the output of the $1D$ convolutional layer. 
Furthermore, to enable effective learning of global representations for the input time series, we prepend a special token \texttt{[start]} to the beginning of the sequence.

\forceindent Next, the input sequence passes through four consecutive Transformer blocks, denoted as \mybox{\texttt{TBlock,8,64${\to}$64}}.
In this notation, the number 8 refers to the attention heads, and the two 64 values represent the input and output dimensions, respectively. 
The design of the Transformer block is shown in Fig.~\ref{fig:tblock}. 
From the output sequence generated by the Transformer blocks, we extract the intermediate representation associated with the \texttt{[start]} token. 
This extracted representation serves as the global representation of the input time series, capturing the essential information from the entire sequence.
This mechanism shares similarities with the \texttt{[CLS]} token used in prior Transformer models like BERT~\cite{devlin2018bert}, highlighting its significance in capturing global context. 
Finally, the output of this backbone model is a size 128 vector that represents each input time series.

\end{itemize}

The detailed design of the residual block can be found in Fig.~\ref{fig:rblock}. 
This block consists of two passages: the main passage and the skip connection passage. 
The main passage processes the input sequence using three pairs of $1D$ convolutional-ReLU layers. 
The convolutional layers have filter sizes of seven, five, and three, sequentially, progressing from the input to the output.
On the other hand, the skip connection passage may include an optional $1D$ convolutional layer with a filter size of one. 
This convolutional layer is only introduced to the skip connection when the input dimension and the output dimension of the residual block differ.
The output of the main passage and the skip connection passage are combined through element-wise addition to form the final output of the block. 

The design of the Transformer block is illustrated in Fig.~\ref{fig:tblock}, showcasing its structure and components. 
The Transformer block is composed of two stages: a multihead self-attention stage and a feed-forward stage.
In the first stage, a multihead self-attention module is employed. 
This module allows the Transformer block to capture dependencies between different positions in the input sequence.
The second stage involves a position-wise feed-forward network. 
This network applies two linear layers with a ReLU activation function to each position of the sequence obtained from the multihead self-attention module.
This stage enables the Transformer block to incorporate non-linear transformations and enhance the representation of each position.
Both stages incorporate skip connections, ensuring that the input is added to the output at each stage. 
This mechanism facilitates the flow of information from the input to the output, enabling the model to retain important information throughout the block.
The Transformer block is responsible for modeling complex dependencies and enhancing the representation of the input sequence through self-attention and position-wise transformations.

We adopt layer normalization~\cite{ba2016layer} for all normalization layers in our model. 
Layer normalization has proven to be effective and is commonly used with sequential data~\cite{ba2016layer,vaswani2017attention}. 
By applying layer normalization, we can ensure stable and consistent normalization across different layers.

\subsection{Model Training and Inference}
The training procedure for the proposed ego-network Transformer model is outlined in Algorithm~\ref{alg:train}. 
The algorithm takes the following inputs: the training time series $T$, the ground truth labels $Y$, the subsequence length $m$, and the number of neighbors $k$.

\begin{algorithm}[ht]
    \centering
    \caption{Training Algorithm\label{alg:train}}
    \footnotesize
    \begin{algorithmic}[1]
        \Input{time series~$T \in \mathbb{R}^n$, ground truth label~$Y \in \mathbb{N}^n$, subsequence length~$m \in \mathbb{N}$, number of neighbors~$k \in \mathbb{N}$}
        \Output{ego-network Transformer model~$f_\theta$}
        \Function{Train}{$T, Y, m, k$}
        \State $\mathcal{G} \gets \textsc{$k$NNSTOMP}(T, m, k)$
        \For{\textbf{each epoch}}
        \State $n_\text{sample} \gets \lceil \frac{n}{m} \rceil$
        \State $\mathbf{X}_\text{sample}, Y_\text{sample}, I_\text{sample} \gets \textsc{SampleSub}(T, Y, n_\text{sample})$
        \For{\textbf{each iteration}}
        \State $\mathbf{X}_\text{batch}, Y_\text{batch}, I_\text{batch} \gets \textsc{GetBatch}(\mathbf{X}_\text{sample}, Y_\text{sample}, I_\text{sample})$
        \State $\mathbf{I}_\text{neighbor} \gets \textsc{GetNeighborIndex}(\mathcal{G}, I_\text{batch})$
        \State $\mathbf{X}_\text{neighbor}, \mathbf{Y}_\text{neighbor} \gets \textsc{GetNeighbor}(T, Y, \mathbf{I}_\text{neighbor})$
        \State $f_\theta \gets \textsc{Update$\theta$}(f_\theta, \mathbf{X}_\text{batch}, Y_\text{batch}, \mathbf{X}_\text{neighbor}, \mathbf{Y}_\text{neighbor})$
        \EndFor
        \EndFor
        \State \Return $f_\theta$
        \EndFunction
    \end{algorithmic}
\end{algorithm}

To begin, in line 2, Algorithm~\ref{alg:knn_stomp} is employed to construct the $k$-nearest neighbor graph $\mathcal{G}$.
Next, in line 4, the number of samples $n_\text{sample}$ is calculated. 
This value corresponds to the minimal number of subsequences used for training in each epoch and is determined based on the number of subsequences required to cover the time series $T$.
Using all subsequences would lead to redundancy due to overlap between subsequences.
In line 5, $n_\text{sample}$ subsequences are randomly sampled from $T$ without replacement.
The sampling process yields the sampled subsequences $\mathbf{X}_\text{sample}$, their associated ground truth labels $Y_\text{sample}$, and the indices of the subsequences $I_\text{sample}$.
This random sampling approach enables the model to train on different shifts of essentially the same subsequences, thereby enhancing its robustness.

In line 7, the mini-batch for the iteration is prepared, consisting of the subsequence $\mathbf{X}_\text{batch}$, the corresponding ground truth labels $Y_\text{batch}$, and the indices $I_\text{batch}$.
In line 8, by utilizing the $k$-nearest neighbor graph $\mathcal{G}$ and the indices $I_\text{batch}$, we obtain the indices of the $k$-neighbors, denoted as $\mathbf{I}_\text{neighbor}$, for each sample in the mini-batch.
It is important to note that $\mathbf{I}_\text{neighbor}$ is a matrix of size $n_\text{batch} \times k$, where $\mathbf{I}_\text{neighbor}[i, j]$ contains the index of the $j$th neighbor for the $i$th sample in the mini-batch.
In line 9, we extract the subsequences and labels for each neighbor based on the indices in $\mathbf{I}_\text{neighbor}$.
Line 10 involves updating the parameters $\theta$ of the ego-network Transformer model $f_\theta$.
Finally, in line 11, the trained model $f_\theta$ is returned as the output of the algorithm.

The inference procedure for the proposed Transformer model is presented in Algorithm~\ref{alg:inference}.
The algorithm accepts the following inputs: the test time series $T_\text{test}$, the training time series $T_\text{train}$, the training labels $Y_\text{train}$, the subsequence length $m$, and the number of neighbors $k$.

\begin{algorithm}[ht]
    \centering
    \caption{Inference Algorithm\label{alg:inference}}
    \footnotesize
    \begin{algorithmic}[1]
        \Input{test time series~$T_\text{test} \in \mathbb{R}^{n_\text{test}}$, training time series~$T_\text{train} \in \mathbb{R}^{n_\text{train}}$, training label~$Y_\text{train} \in \mathbb{N}^n$, subsequence length~$m \in \mathbb{N}$, number of neighbors~$k \in \mathbb{N}$}
        \Output{predicted label~$\hat{Y}_\text{test}$}
        \Function{Inference}{$T_\text{test}, T_\text{train}, Y_\text{train}, m, k$}
        \State $\mathcal{G} \gets \textsc{$k$NNSTOMP}(T_\text{test}, T_\text{train}, m, k)$
        \For{\textbf{each} $\mathbf{X}_i, i \gets \textsc{GetSub}(T_\text{test}, m)$}
        \State $I_\text{neighbor} \gets \textsc{GetNeighborIndex}(\mathcal{G}, i)$
        \State $\mathbf{X}_\text{neighbor}, Y_\text{neighbor} \gets \textsc{GetNeighbor}(T, Y, I_\text{neighbor})$
        \State $\hat{Y}_\text{test}[i] \gets f_\theta(\mathbf{X}_i, \mathbf{X}_\text{neighbor}, Y_\text{neighbor})$
        \EndFor
        \State \Return $\hat{Y}_\text{test}$
        \EndFunction
    \end{algorithmic}
\end{algorithm}

Similar to Algorithm~\ref{alg:train}, Algorithm~\ref{alg:inference} also constructs the $k$-nearest neighbor graph $\mathcal{G}$ in line 2, but with the difference that it finds the $k$-nearest neighbors from $T_\text{train}$ for each subsequence in $T_\text{test}$.
From line 3 to line 6, the algorithm predicts the label for each subsequence $\mathbf{X}_i \in T_\text{test}$, where $i$ denotes the index associated with $\mathbf{X}_i$.
In line 4, the index of the neighbors for the subsequence $\mathbf{X}_i$ is extracted from $\mathcal{G}$.
In line 5, the subsequence and label for each neighbor are extracted from $T_\text{train}$ and $Y_\text{train}$.
In line 6, the label for $\mathbf{X}_i$ is predicted and stored in $\hat{Y}_\text{test}[i]$.
The predicted labels $\hat{Y}_\text{test}$ are returned in line 7.

It's worth noting that although we introduced both Algorithm~\ref{alg:train} and Algorithm~\ref{alg:inference} with univariate time series, both algorithms can also handle multivariate time series.

\subsection{Temporal Consistency Post-processing}
To ensure temporal consistency, we employ a sliding window technique to smooth the predicted label series. 
This post-processing method is both simple and effective. 
By following these steps, we obtain the smoothed label vector $Y_\text{after}$ from the input label vector $Y_\text{before}$ and a sliding window length $m$:

\begin{enumerate}
    \item Starting from the left and progressing towards the right, we identify the next onset for the relevant classes (i.e., non-background classes) in $Y_\text{before}$.
    \item If an onset is found at index $i_\text{onset}$, we determine the majority class within $Y_\text{before}[i_\text{onset}:i_\text{onset} + m]$.
    \item If the majority class identified in the previous step is denoted as $c$, we assign $c$ to $Y_\text{after}[i_\text{onset}:i_\text{onset} + m]$.
    \item Repeat steps 1--3 until reaching the end.
\end{enumerate}

\begin{figure}[ht]
\centerline{
\includegraphics[width=0.8\linewidth]{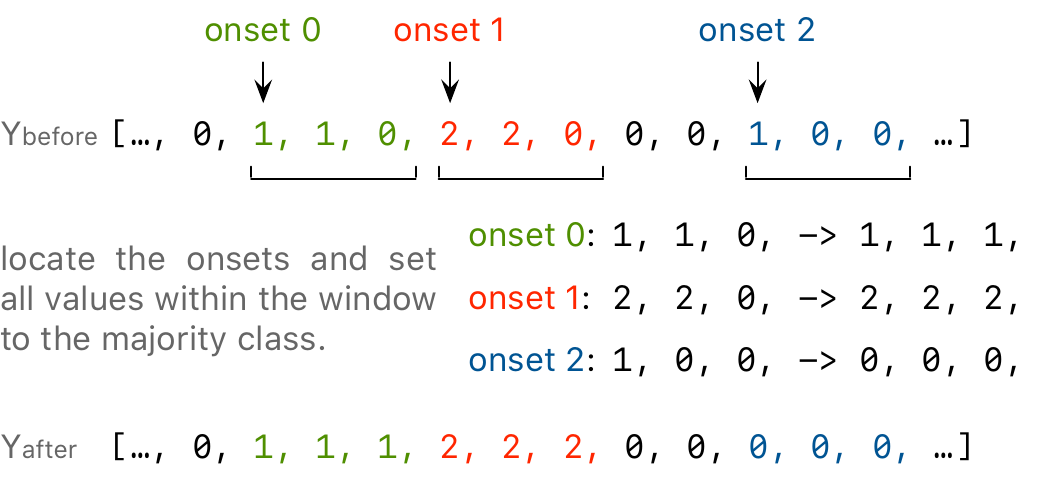}
}
\caption{
The post-processing method enhances the temporal consistency.
The size of the sliding window is three.
}
\label{fig:smooth}
\end{figure}

Fig.~\ref{fig:smooth} provides a visual example illustrating the post-processing method with $m=3$. 
In this example, three onsets (\texttt{onset 0}, \texttt{onset 1}, and \texttt{onset 2}) are detected, and there are three classes (\texttt{class 0} which represents the background, \texttt{class 1}, and \texttt{class 2}). 
For \texttt{onset 0}, the majority class within the window is \texttt{class 1}, so we set the corresponding values in $Y_\text{after}$ to \texttt{class 1}. 
Similarly, for \texttt{onset 1} and \texttt{onset 2}, we assign the corresponding values in $Y_\text{after}$ as \texttt{class 2} and \texttt{class 0}, respectively. 
This post-processing method improves the temporal consistency of the predicted labels. 
In the given example, the likely erroneous classification of \texttt{class 1} at \texttt{onset 2} is corrected. 
Please note that in our experiments, we determine the length of the sliding window using a validation dataset.

\section{Experiment}
\label{sec:experiment} 
In this section, we present the results on 128 univariate and 30 multivariate time series datasets. 
We begin by discussing the dataset preprocessing steps employed in our experiments. 
Next, we describe the performance metrics used to evaluate the models. 
We then introduce the baseline methods that we compare against.
When presenting the experimental results, we first assess the effectiveness of the temporal consistency post-processing step. 
Next, we demonstrate how the utilization of ego-networks enhances the overall performance. 
Finally, we illustrate the reasons for the superior performance of our proposed method compared to the baseline approaches.
Please refer to~\cite{supplementary} for detailed results and access to the source code.

\subsection{Dataset}
The 128 univariate datasets are from the UCR Archive~\cite{dau2019ucr}, while the 30 multivariate time series datasets are from the UEA Archive~\cite{bagnall2018uea}. 
Originally designed for time series classification (as depicted in Fig.~\ref{fig:problem_0}), we have adapted these datasets to suit our problem setting (as shown in Fig.~\ref{fig:problem_2}) by applying the following pre-processing steps:

\begin{enumerate}
    \item Splitting all instances in the dataset into training, validation, and test sets with a ratio of 6:2:2. 
    Each instance is considered as a foreground segment.
    \item Generating the background segment for each foreground segment using a random walk generator, with its length being twice that of each foreground segment.
    For multivariate time series, the background segment is also multivariate.
    \item Splitting the background segment into two parts at a random position and concatenating each part to the beginning and end of the foreground segment. 
    When connecting the foreground segments with the background segments, we carefully adjust the offset to avoid any noticeable step-shape artifacts around the connection points.
    \item Connecting all the expanded instances within each set (i.e., training, validation, and test sets) into continuous time series. 
    Again, we adjust the offset of each instance to prevent the occurrence of step-shape artifacts.
    \item Creating the ground truth labels for each time series by assigning a class label to a subsequence when 60\% or more of the subsequence consists of the foreground segment from that particular class.
\end{enumerate}

\begin{figure}[ht]
\centerline{
\includegraphics[width=0.85\linewidth]{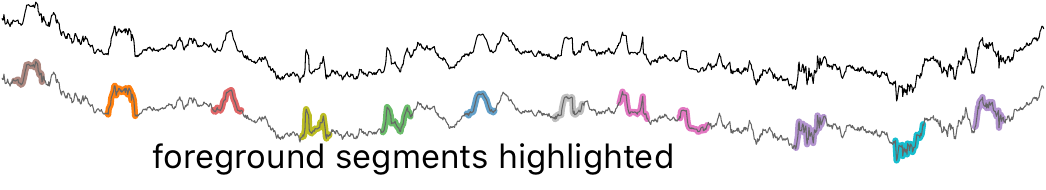}
}
\caption{
The first 1,700 samples from the training time series for the Crop dataset from UCR Archive.
Without the highlighting, it is not easy to identify the foreground segments from the background.
}
\label{fig:example_ts}
\end{figure}

Fig.~\ref{fig:example_ts} illustrates the first 1,700 training time series samples from the Crop dataset in UCR Archive. 
In Fig.~\ref{fig:example_ts}.top, the foreground segments are not highlighted, making it challenging to visually distinguish them from the background.
This demonstrates the difficulty of the subsequence classification task in the presence of background segments.

\subsection{Performance Measure}
Because detecting the onset of an event (i.e., when a foreground segment has just occurred) is more important, we use the onset-based $F1$-score to measure performance. 
The difference from the traditional $F1$-score lies in how precision and recall are calculated.
To calculate precision and recall, we first identify all the onsets (i.e., the beginning of a foreground class) in the predicted label $\hat{Y}_\text{test}$ and the ground truth label $Y_\text{test}$. 
For each onset in $\hat{Y}_\text{test}$, we check if the onset location contains a ``correct" prediction or not. 
If the total number of onsets in $\hat{Y}_\text{test}$ is denoted as $n_\text{total}$ and the number of correctly predicted onsets as $n_\text{correct}$, the precision is computed as $\frac{n_\text{correct}}{n_\text{total}}$.
We define a correct prediction as follows: Given an onset location $i$ in $\hat{Y}_\text{test}$, if there exists a location $j$ in $Y_\text{test}$ such that $\hat{Y}_\text{test}[i] = Y_\text{test}[j]$ and $|i - j| < 0.1m$, where $m$ is the length of the median foreground segment.
To compute the recall, we repeat the above steps by swapping the roles of $\hat{Y}_\text{test}$ and $Y_\text{test}$.

The experiments were conducted on multiple datasets, and in order to compare the overall performance of each method, it is necessary to summarize the performances across these datasets.
For each compared method, we determined its ranking based on the $F1$-score achieved on each individual dataset. 
By averaging these rankings, we obtained the overall ranking for each method, which we report in this paper.
Please note that due to space limitations, we only present the summarized results. 
For readers interested in the complete results, we refer them to~\cite{supplementary}.

\subsection{Baseline Methods}
There are two sets of baselines that we compared in this experiment.
The first set of methods consists of $k$-nearest neighbor classifiers with different values of $k$. 
Specifically, we have the one-nearest neighbor (1NN), five-nearest neighbor (5NN), ten-nearest neighbor (10NN), and $k$-nearest neighbor where $k$ is determined using each dataset's validation set.
We use $z$-normalized Euclidean distance with the $k$-nearest neighbor classifiers.
The second set of baseline methods are neural network-based methods, where we add a classification layer on top of each backbone model (i.e., LSTM, GRU, ResNet, and Transformer). 
These baselines are used to highlight the benefits of the proposed ego-network Transformer model.

\subsection{Experiment Result}
The experimental results for UCR Archive and UEA Archive are presented in Table~\ref{tab:result_ucr} and Table~\ref{tab:result_uea}, respectively.
In each table, the first four rows contain the performance of the $k$-nearest neighbor methods, with and without the post-processing technique.
Rows five to eight display the performance of the neural network-based methods.
These include the baseline neural network methods with various backbone models, with and without the post-processing technique, as well as the ego-network Transformer with different backbone models, also with and without the post-processing technique.
For the ego-network Transformer results, we provide the performance for both the five-nearest neighbor graph and the ten-nearest neighbor graph.
The best method is highlighted in bold, while the second best is underlined in the tables.

\begin{table}[ht]
\caption{Experiment results\protect\footnotemark~from UCR Archive.}
\begin{center}
\resizebox{0.95\columnwidth}{!}{
\begin{tabular}{l||c|cc|c|cc}
\multirow{3}{*}{Avg. rank ($\downarrow$)} & \multicolumn{3}{c|}{Not post-processed} & \multicolumn{3}{c}{Post-processed} \\ \cline{2-7}
~ & \multirow{2}{*}{Baseline} & \multicolumn{2}{c|}{Ego-Network} & \multirow{2}{*}{Baseline} & \multicolumn{2}{c}{Ego-Network} \\
~ & ~ & 5NN & 10NN & ~ & 5NN & 10NN  \\ \hline \hline
1NN & 28.54 & - & - & 17.43 & - & -  \\ 
5NN & 26.36 & - & - & 18.79 & - & -  \\ 
10NN & 27.29 & - & - & 21.10 & - & -  \\ 
$k$NN & 26.77 & - & - & 17.74 & - & -  \\ \hline
LSTM & 26.66 & 20.80 & 21.03 & 23.32 & 13.25 & 13.54  \\
GRU & 21.85 & 20.21 & 19.77 & 18.00 & 12.47 & 13.02  \\ 
ResNet & 8.31 & 8.88 & 9.01 & 7.00 & \textbf{5.79} & \underline{5.95}  \\ 
Transformer & 16.63 & 15.08 & 14.46 & 12.38 & 8.21 & 8.36 \\
\end{tabular}
}
\label{tab:result_ucr}
\end{center}
\end{table}

\footnotetext{This table contains 32 methods, each of them will receive a ranking score between 1-32 for a single dataset. 
We average the ranking scores for each method across all datasets and use this average ranking score to evaluate them. 
The smaller this average score is, the better performance the corresponding method has.}

\begin{table}[ht]
\caption{Experiment results$^1$ from UEA Archive.}
\begin{center}
\resizebox{0.95\columnwidth}{!}{
\begin{tabular}{l||c|cc|c|cc}
\multirow{3}{*}{Avg. rank ($\downarrow$)} & \multicolumn{3}{c|}{Not post-processed} & \multicolumn{3}{c}{Post-processed} \\ \cline{2-7}
~ & \multirow{2}{*}{Baseline} & \multicolumn{2}{c|}{Ego-Network} & \multirow{2}{*}{Baseline} & \multicolumn{2}{c}{Ego-Network} \\
~ & ~ & 5NN & 10NN & ~ & 5NN & 10NN  \\ \hline \hline
1NN & 26.67 & - & - & 22.08 & - & -  \\ 
5NN & 26.03 & - & - & 21.97 & - & -  \\ 
10NN & 25.43 & - & - & 22.03 & - & -  \\ 
$k$NN & 25.50 & - & - & 21.67 & - & -  \\ \hline
LSTM & 21.37 & 18.88 & 19.18 & 17.10 & 12.82 & 13.20  \\ 
GRU & 21.15 & 19.90 & 19.83 & 16.67 & 14.57 & 13.58  \\ 
ResNet & 12.17 & 9.02 & 11.10 & 9.98 & 8.85 & \underline{8.50}  \\ 
Transformer & 15.67 & 14.07 & 12.87 & 10.47 & 8.72 & \textbf{6.97} \\ 
\end{tabular}
}
\label{tab:result_uea}
\end{center}
\end{table}

First, we compare the performance of each method with and without the post-processing technique.
It is evident that the post-processing technique enhances the performance of \textit{all} methods.
This implies that the post-processing technique is a simple yet effective method for improving the performance of most subsequence classification systems.

Next, let us compare the baseline $k$-nearest neighbor methods with the baseline neural network methods.
In the UCR Archive (i.e., Table~\ref{tab:result_ucr}), LSTM and GRU exhibit similar performance to the $k$-nearest neighbor baselines, while ResNet and Transformer outperform the $k$-nearest neighbor baselines.
In the UEA Archive (i.e., Table~\ref{tab:result_uea}), all neural network methods surpass the performance of the $k$-nearest neighbor baselines.
One possible explanation for this observation is that the UEA Archive comprises multivariate time series data.
Even with less effective architectures (such as LSTM and GRU), the neural network methods still have an advantage due to their ability to learn the relative importance of different dimensions.

Lastly, we compare the ego-network Transformer methods with their baseline counterparts.
The ego-network Transformer consistently improves performance in almost all cases, whether it utilizes a five-nearest neighbor graph or a ten-nearest neighbor graph.
The only exception is when ResNet is used as the backbone model without the post-processing step in the UCR Archive (see Table~\ref{tab:result_ucr}, row seven, first three columns).
Nevertheless, the best performing method in both tables utilizes the proposed ego-network Transformer (with either the ResNet or Transformer backbones) along with the post-processing step.

\subsection{Case study: Traffic Time Series}
To gain insights into the performance improvement brought by the proposed ego-network Transformer model in subsequence classification, we conducted a detailed analysis using the Dodgers Loop Sensor dataset~\cite{ihler2006adaptive}.
This dataset consists of time series data generated by monitoring highway traffic near the Dodgers Stadium, and the task is to detect whether a baseball game is being hosted at the stadium or not.
It is worth noting that this is a challenging dataset since the Dodgers Stadium also hosts other events like concerts~\cite{wikidodgers}, which may result in a similar amount of traffic around the stadium (i.e., false positives).

The length of the time series is 50,400, and we split it into training, validation, and test sets with a ratio of 6:2:2. 
After the split, there are 46 games in the training time series, 18 games in the validation time series, and 17 games in the test time series.
We utilize the onset-based $F1$-score as the performance metric, and we always apply the post-processing technique to ensure temporal consistency. 
The experiment results are presented in Table~\ref{tab:dodgers}, where the best performance is indicated in bold and the second best performance is underlined.

\begin{table}[ht]
\caption{Utilizing ego-networks improves performance.}
\begin{center}
\resizebox{0.99\columnwidth}{!}{
\begin{tabular}{l||ccc|cccc}
$F1$-score ($\uparrow$) & 1NN & 2NN & 3NN & LSTM & GRU & ResNet & Transformer  \\ \hline \hline
Baseline & 0.15 & 0.15 & 0.00 & 0.00 & 0.00 & 0.00 & 0.18  \\ 
Ego-Network & - & - & - & \underline{0.32} & 0.29 & 0.20 & \textbf{0.37} \\ 
\end{tabular}
}
\label{tab:dodgers}
\end{center}
\end{table}

First, we focus on the performance of the $k$-nearest neighbor method with different values of $k$. 
We observe that the performance for $k=1$ and $k=2$ is identical (i.e., 0.15), while setting $k=3$ reduces the $F1$-score to zero. 
This finding indicates that the top two nearest neighbors play a more important role in determining the class of each test subsequence.
Next, we examine the performance of different baseline backbone neural network methods, which mostly exhibit poor performance with zero $F1$-score. 
The only exception is the Transformer baseline, which outperforms the $k$-nearest neighbor classifier slightly. 
We suspect that their poor performance is due to the limited number of training examples for the positive class. 
However, when we combine each backbone model with the proposed ego-network Transformer, their performance improves dramatically. 
The incorporation of nearest neighbor information mitigates the data scarcity issue associated with this dataset and has the potential to be extended to other types of data.

The overall best model for this dataset is the combination of the ego-network Transformer with the Transformer backbone model. 
Therefore, we further investigate the relative importance of each nearest neighbor in the ego-network. 
To conduct this study, we evaluate the model after removing the $i$th nearest neighbor. 
Removing the first nearest neighbor results in a 72\% reduction in performance, yielding an $F1$-score of 0.10. 
Removing the second nearest neighbor reduces the performance by 49\% to an $F1$-score of 0.19. 
Finally, removing the third nearest neighbor leads to a 32\% performance reduction, with an $F1$-score of 0.25. 
These results further confirm our assumption regarding the importance of the nearest neighbor graph when working with subsequences. 
Notably, these observations align with previous works in the Matrix Profile literature~\cite{yeh2018time,yeh2018towards,zhu2020swiss}.

\section{Conclusion}
\label{sec:conclusion}
In this paper, we present a novel ego-network Transformer model specifically designed to tackle the subsequence classification problem. 
Through extensive experiments on 128 univariate and 30 multivariate time series datasets, we demonstrate the superior performance of our proposed model compared to the baselines.
Furthermore, our in-depth analysis of the proposed method reveals its remarkable effectiveness in addressing data scarcity issues commonly encountered in subsequence classification tasks. 
This finding underscores the model's ability to leverage the nearest neighbor graph and overcome limited training examples for foreground classes.
Overall, our study not only showcases the superior performance of the ego-network Transformer model but also provides empirical evidence supporting the significance of the nearest neighbor graph in subsequence analysis.
For future work, we could consider adopting the novel ResNet$2D$ design~\cite{yeh2023efficient,yeh2023temporal,yeh2023multitask}, pretraining methods~\cite{ma2023survey,yeh2023toward}, tackling scalability issues~\cite{yeh2022embedding,yeh2023sketching}, or addressing data privacy issues~\cite{yeh2023time} for subsequence classification.


\Urlmuskip=0mu plus 1mu\relax
\bibliographystyle{IEEEtran}
\bibliography{section/reference.bib}

\end{document}